\documentclass[letterpaper, 10 pt, journal, twoside]{IEEEtran}
\usepackage[T1]{fontenc}
\usepackage[latin9]{inputenc}
\usepackage{color}
\usepackage{verbatim}
\usepackage{float}
\usepackage{amsmath}

\usepackage{amsthm}
\usepackage{amssymb}
\usepackage{stmaryrd}
\usepackage{stackrel}
\usepackage{graphicx}
\usepackage{wasysym}
\usepackage{mathtools}
\usepackage{hyperref}
\usepackage{url}   
\usepackage{booktabs} 
\usepackage{nicefrac} 
\usepackage{authblk}
\usepackage{multirow}
\usepackage{xcolor}
\usepackage{microtype}
\usepackage{paralist}
\usepackage{paralist}
\usepackage{subfig}
\makeatletter

\DeclareMathOperator*{\argmax}{arg\,max}

\floatstyle{ruled}
\newfloat{algorithm}{tbp}{loa}
\providecommand{\algorithmname}{Algorithm}
\floatname{algorithm}{\protect\algorithmname}

\theoremstyle{plain}

\theoremstyle{definition}

\theoremstyle{definition}

\theoremstyle{plain}

\theoremstyle{definition}

\theoremstyle{remark}

\theoremstyle{plain}

\usepackage{cite}
\usepackage{algpseudocode}
\usepackage{setspace}
\IEEEoverridecommandlockouts

\makeatother

\usepackage{babel}
\providecommand{\corollaryname}{Corollary}
\providecommand{\definitionname}{Definition}
\providecommand{\examplename}{Example}
\providecommand{\lemmaname}{Lemma}
\providecommand{\problemname}{Problem}
\providecommand{\remarkname}{Remark}
\providecommand{\theoremname}{Theorem}

\usepackage{amsmath}               
  {
      \theoremstyle{plain}
      
      \theoremstyle{plain}
      
  }
  
\SetSymbolFont{stmry}{bold}{U}{stmry}{m}{n}

\def\BibTeX{{\rm B\kern-.05em{\sc i\kern-.025em b}\kern-.08em
		T\kern-.1667em\lower.7ex\hbox{E}\kern-.125emX}}
		
\begin{document}

\title{Hierarchical Deep Learning for Intention Estimation of Teleoperation Manipulation in Assembly Tasks

\thanks{$^{1}$Mingyu Cai is with the Department of Mechanical Engineering, University of California, Riverside, CA, USA, 92521. This work is done when he was working with Honda Research Institute.
Karankumar Patel, Soshi Iba, Songpo Li are with Honda Research Institute, San Jose, CA, 95134 USA. $^{*}$ Both authors contributed equally to this research.}
}

\author{$^{*}$Mingyu Cai, $^{*}$Karankumar Patel, Soshi Iba, Songpo Li
}

\markboth{2024 IEEE International Conference on
Robotics and Automation (ICRA), Preprint Version.}
{Cai \MakeLowercase{\textit{et al.}}: Hierarchical Deep Learning for Intention Estimation of Teleoperation Manipulation in Assembly Tasks} 

\maketitle

\begin{abstract}
In human-robot collaboration, shared control presents an opportunity to teleoperate robotic manipulation to improve the efficiency of manufacturing and assembly processes. Robots are expected to assist in executing the user's intentions. 
To this end, robust and prompt intention estimation
is needed, relying on behavioral observations. The framework presents
an intention estimation technique at hierarchical levels i.e., low-level actions and high-level tasks, by incorporating multi-scale hierarchical information in neural networks. Technically, we employ hierarchical dependency loss to boost overall accuracy. Furthermore, we propose a multi-window method that assigns proper hierarchical prediction windows of input data. 
An analysis of the predictive power with various inputs demonstrates the predominance of the deep hierarchical model in the sense of prediction accuracy and early intention identification. We implement the algorithm on a virtual reality (VR) setup to teleoperate robotic hands in a simulation with various assembly tasks to show the effectiveness of online estimation.
Video demonstration is available at: 
\url{https://youtu.be/CMYDgcI4j1g}.
\end{abstract}



\section{INTRODUCTION}

Shared autonomy to enable close human-robot collaboration is being actively investigated in industrial applications and surgical tasks~\cite{alevizos2020physical, fang2023human, wilcox2013optimization, zhou2023local}.
Teaming up humans' dexterity and mechanic capability of robots boosts production efficiency, raising the need for robotic teleoperation.
Whenever a flexible and skilled
manual action is required without access to human's physical presence, teleoperation could provide a means to remedy
the situation~\cite{li2022assimilation}. It involves a wide range of applications e.g., healthcare to safely provide medical assistance to contagious patients, industrial productions requiring sterile environments, and assistive applications restoring arm mobility to impaired users.

However, it's still challenging to operate a robot for non-experts since perception and action are in this case both mediated by technical systems, they are also
possibly affected by delays.
For seamless physical human-robot collaboration, the robot has to understand
human performance and intentions to be able to provide effective
and transparent assistance~\cite{schydlo2018anticipation, nicolis2018human, belardinelli2022intention, huang2023hierarchical}.
This work focuses on reliable human intention estimation for assistive motion control which is a critical
component of safe and seamless robot teleoperation. 

Existing works of human intention estimation investigate either grasping goals~\cite{nicolis2018human, belardinelli2022intention, manschitzshared} or analyzing single short-horizon actions\cite{schydlo2018anticipation}. However, they
failed to fully reason about the contextual relations between
adjacent actions under an umbrella of one certain structured task, which provide potential temporal logic for understanding long-term intention and prediction. For instance, when a human-robot team is placing a screw on a wheel, there is a big probability of taking a screwdriver as the next action and the assembling target is likely to be an auto toy. Moreover, it is also necessary to know the task to generate the proper assistance to the current action. For example, the grasping constraints of the same object would be different in an usage task and for a relocation task. 

 \begin{figure}
	\centering 	\includegraphics[width=0.35\textheight]{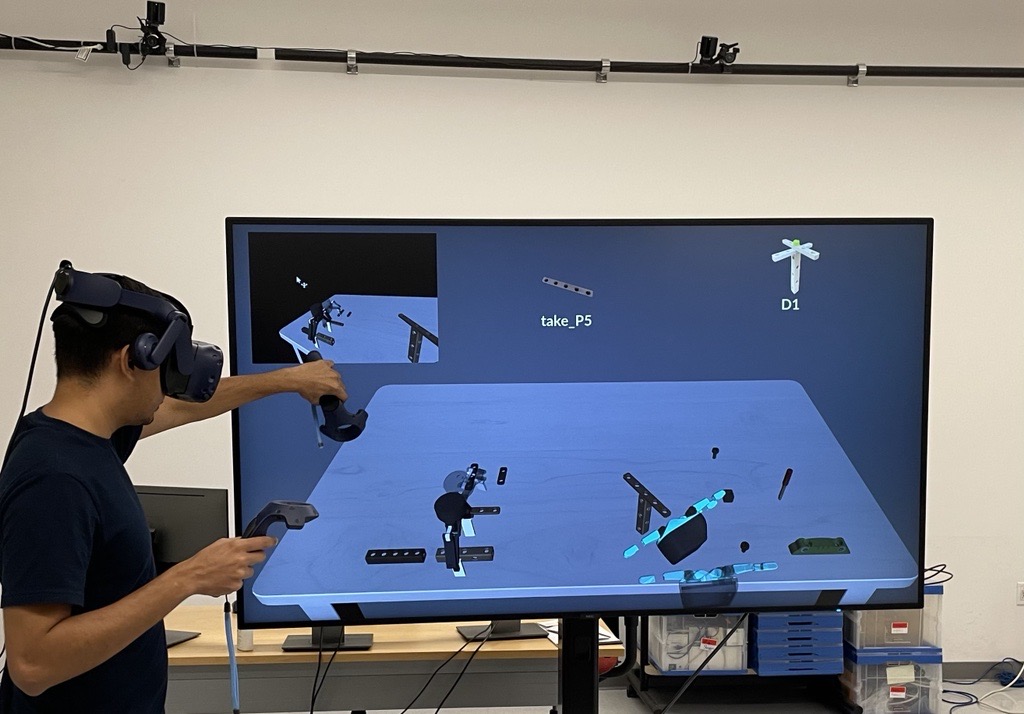}
	\caption{Experimental setup for the data collection and model testing. 
 The movements of human operator's head, hands, and eye gaze are tracked via HTC Vive virtual reality system. 
 The top-left corner of screen visualizes the scene as perceived by the operator's point of view, and the background scene shows the global view of a teleoperation process. Action and task estimation results are shown in the middle and top right screen respectively.}
	\label{demo}
        \vspace{-3mm}
\end{figure}

\textbf{Contribution }
In this work, we formulate intention estimation at hierarchical levels.
In particular, the low-level intention estimation tracks fine actions for control assistance. The high-level mechanism is to predict human's long-horizon coarse tasks, which provides useful instructions of action sequences. 
Instead of developing separate models for each level that may cause hierarchical inconsistency, we are inspired by the hierarchical classification strategy~\cite{gao2020deep} and extend it to the sequential neural network models. The novelty is to
incorporate dependency information of hierarchical layers in a top-down manner, where the output of the lower level is conditioned by its upper level. We present three main contributions:

\begin{compactitem}
    \item Different from previous method~\cite{gao2020deep}, our hierarchical levels require different sequential lengths of data, resulting inconsistent multi-input horizons. We address this issue by proposing a multi-window strategy that forwards the input data with a different range of masks to achieve flexible hierarchical data inputs. 

    \item Compared with the standard method, we show that the layer-dependent deep hierarchical model is capable of improving the estimation performance using inputs from either motion data or visual egocentric view data.

    \item A new assembly dataset was collected in a virtual reality setup with two robot hands in a simulation to manipulate objects in teleoperation. The online performance is demonstrated through $6$ assembly tasks with $21$ actions in total.
\end{compactitem}

It's also worth pointing out that our architecture can be easily extended with the state-of-the-art estimation models for more sophisticated intentions and performance improvement.

\vspace{0.2cm}
\textbf{Related Works }
In the context of teleoperation, advancing
autonomy mainly addresses two challenges: predicting the operator's intent in performing a task and deciding how to assist the teleoperator~\cite{sheridan1995teleoperation, rozo2016learning}.
Existing literature in general describes the what-to-predict and how-to-assist problems . After inferring the operator's intentions, many works~\cite{yu2005telemanipulation, dragan2013policy, hauser2013recognition} integrated cooperative motion planners and learning-based policies from demonstrations.
Within the concept of human intention estimation, 
early approaches and several recent ones formulated the user
control input in driving the robotic movement as behavioral
cue for inference and prediction~\cite{aarno2008motion, tanwani2017generative}.
To predict a
distribution over the different action targets, most of these works fused robot motion features such as end effector
pose, velocity, arm joints, or whole gestures, and various types of observation
on human behavior including human
trajectories,  gesture, gaze information giving Area-of-Interest of teleoperators, speech, facial expressions, and force-torque measurements. In this paper, we focus on intetions recognitions for high-level actions and takks.

Along the line of intention estimation, Hidden Markov Models
(HMMs) have been widely used to analyze a discrete set of tasks/subtasks~\cite{rabiner1989tutorial, tanwani2016learning, tanwani2017generative}. These works are studied on a single-layer, whereas human intention is often composed of a multilayer hierarchy.
The use of hierarchical HMM representations has been investigated for multi-layer classifications~\cite{aarno2008motion, zhu2008human, huang2023hierarchical}.
The aforementioned works generally infer the probability distribution over intentions by dynamic programming, which may be computational expensive for online performance with rich and long sequential observations, and also increases the complexity of modeling.  
Neural networks (NNs) have seen increasing popularity in robotics, and sequential NN models e.g., RNN and transformer, are becoming powerful tools for human-robot situation understanding~\cite{li2013human, de2016neural, lea2016learning, nicolis2018human, yuan2022leveraging,lu2020human}. These works detect user motion intent from limb dynamics and from various sensors to enforce collaboration tasks. Direct feed-forward after training makes NNs efficient in practice. 
The hierarchical structure of intentions in human-robot interaction has not been thoroughly explored in neural networks literature, and this study compares its accuracy within this context.
 A command can be interpreted as a pyramid of a goal, sub-goals, and primitives.
Only a few existing works~\cite{bandouch2009tracking, hayes2016autonomously, holtzen2016inferring, han2016interactive}  designed a hierarchical network based on topological properties of graphical task representations. However, their models didn't include the top-down relation during training and required experts to pre-construct the graph structure.


\vspace{-0.15cm}
\section{Problem Formulation}

A collaboration team of human teleoperating robots is assigned a set of $m$ toy assembly tasks denoted as $T$, which
aim to build desired targets e.g., airplanes, vehicles, and block buildings. 
The human teleoperator attempts to take a set of $n$ actions in total denoted as $A$ e.g., pick up a screwdriver, screw track with the left hand, pick up a toy block, etc, and actively leads the team to complete all tasks by performing actions sequences that are unknown to the robot. We define the human intention at time-step $t$ as $H_{t} = (T_{t}, A_{t})$, where $T_{t}\in T$ and $A_{t}\in A$ represent task and action attempted to perform at time $t$. 
With modern sensor equipment, the online observations history $X_{1:t}\in \mathbb{R}^{t\times F}$ is available that includes information on intention estimation e.g., human-robot motion features,  videos of surrounding cameras, egocentric views, gaze, etc, where $F$ denotes the number of input features.
To achieve seamless teleoperation, it's expected to online capture the intention of the teleoperator and subsequently provide autonomous shared control as assistance. 

\begin{figure}
	\centering 	\includegraphics[width=0.35\textheight]{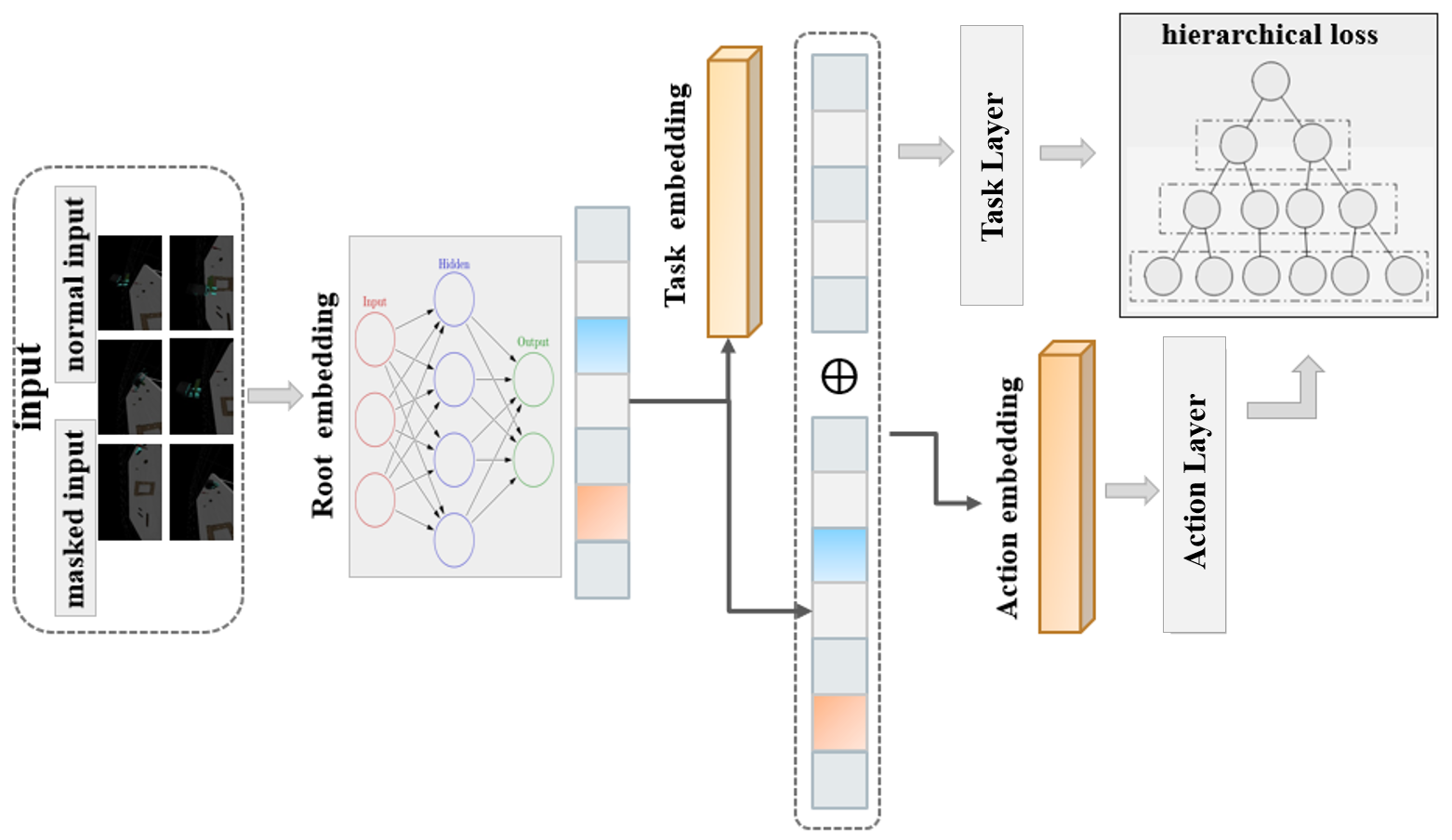}
	\caption{The task-action hierarchical deep learning model including dependent loss functions and leaf layers conditional by the embeddings from its root layer.}
	\label{Slow_Fast}
\end{figure}

Different from existing works, this work considers the hierarchical intention relations shown in Fig.~\ref{Slow_Fast}. In practice, each task $T_{t}$ does not include all action categories. For instance, the block-building task never involves the actions related to screws. We denote $A^{T}_{t}$ as the set of actions that the task $T_{t}$ only takes from. The problem can be formulated as:
at every time-step $t$ with the observation history $X_{1:t}$, the objective is to efficiently predict the teleoperator's intention $H_{t} = (T_{t}, A_{t})$ with hierarchical relations online s.t., $A_{t}\in A^{T}_{t}$.

\section{Method}

\subsection{Deep Hierarchical Model}

Always taking $X_{1:t}$ as the input results in issues of dynamic input and numerous lengths. Assigning the proper window size for the sequential data is a common modeling technique that is applied to process datasets. We denote $L$ as the selected window size, and our model only considers the most recent $L$ time-steps. Consequently, the dataloader generates $X\in\mathbb{R}^{L\times F} = X_{t-L:t}$ as the input. And the ground truth intention $H_{t} = (T_{t}, A_{t})$ only depends on the attempted behavior at the current time step $t$, which can be generated through standard annotation process through the manual segmentation and labeling of the collected dataset of actions and tasks.~\cite{feichtenhofer2019slowfast}. Annotation efficiency could also be achieved by employing a hybrid approach that combines human labeling with state-of-the-art segmentation models.

The categories are organized by a tree with two hierarchical layers, where task prediction $Y_{T}$ is the root layer of action inference $Y_{A}$. Let $\Tilde{T}$ and $\Tilde{A}$  denote the outputs of layers  $Y_{T}$ and  $Y_{A}$ at time-step $t$. Since the predictions rely on contextual relations of observation history, this framework applies the sequential neural network models as the backbone e.g., RNN~\cite{medsker2001recurrent}, LSTM,~\cite{graves2012long} transformer~\cite{vaswani2017attention}, Slow-Fast~\cite{feichtenhofer2019slowfast} , etc. We define the applied backbone (root) neural network as $\mathcal{N}_{r}(X, \theta_{r})$, where $\theta_{r}$ are the parameters to be trained. Its output can be regarded as the root latent space: $X_{r} = \mathcal{N}_{r}(X, \theta_{r})$.

Given the root representation, the objective is to generate hierarchical representations for task and action layers. Since the action layer is the leaf node of the task layer, we design the neural network structure such that the action prediction is conditioned on the task inference i.e., $P(\Tilde{A}_{t} | \Tilde{T}_{t}))$, where $P(\cdot | \cdot)$ represents the conditioned probability.
To do so, first, we construct the task and action encoders, respectively, i.e., 
$X_{T} = \mathcal{N}_{T}(X_{r}, \theta_{T})$ and $X_{A} = \mathcal{N}_{A}(X_{r}, \theta_{A})$.

Then, the task classification layer can be designed using softmax
regression as 

$$\Tilde{y}_{Ti} =\frac{exp(W_{Ti}*X_{T})}{\sum_{k=1}^{m} exp(W_{Tk}*X_{T})}$$

where $W_{Ti}$ are the parameters (weights) of $i$th task category.
To condition the prediction of action, we first concatenate the action and task embeddings i.e., $X_{A|T} = X_{A} \oplus X_{T}$. Similarly, the action classification layer can be constructed as

$$\Tilde{y}_{Ai} = \frac{exp(W_{Ai}*X_{A|T})}{\sum_{k=1}^{n} exp(W_{Ak}*X_{A|T})}$$

where $W_{Ai}$ are the parameters (weights) of $i$th action category.
Finally, the inference results $\Tilde{T}$ and $\Tilde{A}$ can be obtained by taking the $\argmax$ of $\Tilde{y}_{T}$ and $\Tilde{y}_{A}$.

The classification loss function of action and task is designed through standard classification entropy loss as:

$$ELoss = -T_{t} \cdot \log (\Tilde{T}_{t}) - A_{t} \cdot \log (\Tilde{A}_{t})$$

To enhance the hierarchy relations, we introduce $\mathbb{D}$, $\mathbb{I}_{A}$, and $\mathbb{I}_{T}$ to indicate whether the intention predictions of neural network model have conflict hierarchical category structure i.e., $A_{t}\notin A^{T}_{t}$, especially

$$\mathbb{D}=
\begin{cases}
1 & \text{if } \Tilde{A}_{t}\in A^{T}_{t}\\
0 & \text{otherwise}
\end{cases}
$$

$$\mathbb{I}_{T}=
\begin{cases}
1 & \text{if } \Tilde{T}_{t} = T_{t}\\
0 & \text{otherwise}
\end{cases}
, \mathbb{I}_{A}=
\begin{cases}
1 & \text{if } \Tilde{A}_{t} = A_{t}\\
0 & \text{otherwise}
\end{cases}
$$


Based on that, the hierarchical dependence loss is formulated as:
$$DLoss = -(ploss)^{\mathbb{D}\cdot \mathbb{I}_{A}} * (ploss)^{\mathbb{D}\cdot \mathbb{I}_{T}}.$$

where $ploss$ serves as a penalty that enforces the neural network to acquire structural information from the category arrangement. The value of $ploss$ can either be fixed as a constant or be linked to the prediction error.
The total loss of the model is defined as the weighted
summation of the classification entropy loss $Eloss$ and hierarchical dependence loss $Dloss$ i.e., 
$$Loss(\theta) = \alpha \cdot ELoss + \beta \cdot DLoss,$$
where $\alpha\in (0,1)$, $\beta\in (0, 1)$ are tuning parameters to bias the weights of different loss functions.

\subsection{Multi-window Strategy \label{sub:mask}}

In the practice of sequential models, the length of input data is crucial for classification accuracies. In our case, the action inference needs a shorter length of input sequential data compared with task prediction, whereas many deep learning models require a fixed length of input sequential data. Directly sharing the same input with the longest length of data for both task and action recognition is not ideal, since additional unnecessary information may confuse the action inference model and downgrade its performance.

To address the issue and achieve more informative inputs, we developed the multi-window method using the mask technique. In particular, we create the latent embedding space for task and action, respectively, in the model. Each window takes different inputs such that we make the unnecessary horizons of input data for action embeddings. This allows the model to discard masked information and operate only on useful data horizons at hierarchical levels.

In particular, 
we denote $M\in \left\{0, 1\right\}^{(L_{0}+L_{1})}$ as the sequential mask vector generated by users, where $L_{0}+L_{1} = L$ and $0$ indicates a time-step (index) is invisible for the model, and vice versa. Here $L_{0}$ represents the prefix data that should be masked, and $L_{1}$ represents the suffix data that should be kept the same. 
Let $M[i]$ denote the $i$th element of the mask vector.
Thus, we have $M[i] = 0, \forall L_{0}\geq i>0$, and $M[j] = 1, \forall L_{1}\geq j>L_{0}$. Given the current input $X\in\mathbb{R}^{L\times F}$, the mask process generates the representation of valid inputs $\hat{X}\in\mathbb{R}^{L\times F}$ as

\begin{equation}
\hat{X} = Mask(X, M)= \left\{X[i]: 
\begin{cases}
X[i] & \text{if } M[i]=1\\
0 & \text{otherwise}
\end{cases}
\right\}    
\label{eq:mask}
\end{equation}

In this framework, we choose the $L$ of input data $X$ as the sequential length for parent (task) layer prediction, since the one task has longer horizons including a sequence of actions. Then, $L_{1}$ is selected according to the longest duration of all actions. Finally, we forward $X$ and $\hat{X} = Mask(X, M)$ as the input of task and action embedding models, respectively.
As a result, the task and action embedding can be produced in a heterogeneous way i.e., 
\begin{equation}
   \begin{cases}
    X_{T} = \mathcal{N}_{T}( \mathcal{N}_{r}(X, \theta_{r}), \theta_{T}) \\
    X_{A} = \mathcal{N}_{A}( \mathcal{N}_{r}(Mask(X, M), \theta_{r}), \theta_{A}).
\end{cases} 
\label{eq:embeding}
\end{equation}

\subsection{Vision-based Deep Hierarchical Model}

 \begin{figure}
	\centering 	\includegraphics[width=0.35\textheight]{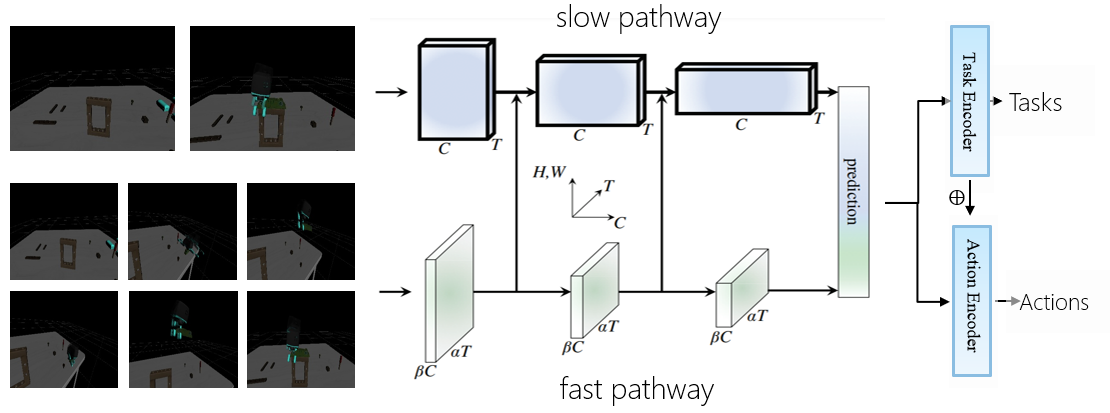}
	\caption{Hierarchical Slow-Fast model accepts only visual inputs without feature extractions.}
	\label{Fig:Slow_Fast}
 \vspace{-1.cm}
\end{figure}

With only visual input, we apply Slow-Fast model~\cite{feichtenhofer2019slowfast}
 that can learn useful temporal information for video recognition, as the backbone to further test the performance of the deep hierarchical model. It includes two pathways i.e., a Slow pathway, operating at a low frame
rate to capture spatial semantics, and a Fast pathway operating at a high frame rate, to capture motion at fine temporal resolution. It has shown SOTA accuracy on popular benchmarks, Kinetics, Charades, and AVA.
The motivation for applying such types of neural network models is that data on motion features may not always be available. In practice, it's desired to predict intentions only using perception information.

In this work, we extend the standard Slow-Fast model by integrating the developed hierarchical structure as shown in Fig.~\ref{Fig:Slow_Fast}, where the primary model extracts temporal and spatial information as the inputs of task and action embeddings. The mask mechanism in section~\ref{sub:mask} is modified in a way that $X[i]\in\mathbb{R}^{H\times W\times C}$ of egocentric history $X\in \mathbb{R}^{L\times H\times W\times C}$ is the frame images, where $H, W, C$ represent with, height, and number of channels. The mask process is still the same as~\eqref{eq:mask}
by setting the elements of the $3D$ matrix $X[i]$ as zero. Finally, we can add a multilayer perception at the end to produce task and action embeddings as~\eqref{eq:embeding}.

\subsection{Manipulation Assistive Control}

Once we obtain the results of intention estimation, we can subsequently pass it into developed assistive control modules to provide autonomous AI support and mitigate the operational workload. There are two common existing formulations to assist the teleoperator.
First, shared control is popular to correct the movement of the robot arm according to intention predictions~\cite{tanwani2017generative, abi2017learning}.

On the other hand, continuously operating the remote arm for routine tasks can be cumbersome for the
teleoperator, especially in the presence of communication
latency. In such a scenario, the teleoperator may relax and the system will automatically switch to autonomous control mode
which the robot arm takes the intention estimation as inputs and recursively re-plans through trajectory generation~\cite{feddema1989vision, marcucci2022motion} or imitation learning policies~\cite{rozo2016learning, shi2023waypoint}, to executes the task and corresponding action sequence for the next steps. The operator can take over the control back at any time and customize their desired behaviors.

When conducting remote teleoperations, the teleoperator only receives perception information from the virtual fixtures of the robots, resulting in the reality gap between the operator's simulated scenarios and the robots' actual workspace. 
In particular, it's challenging for humans to establish real contacts remotely. Our framework includes an AI support module that leverages the manifold information of objects to align the motion of robot end-effector with the desired contact path.

\section{Experimental Results}

 \begin{figure}
	\centering 	\includegraphics[width=0.35\textheight]{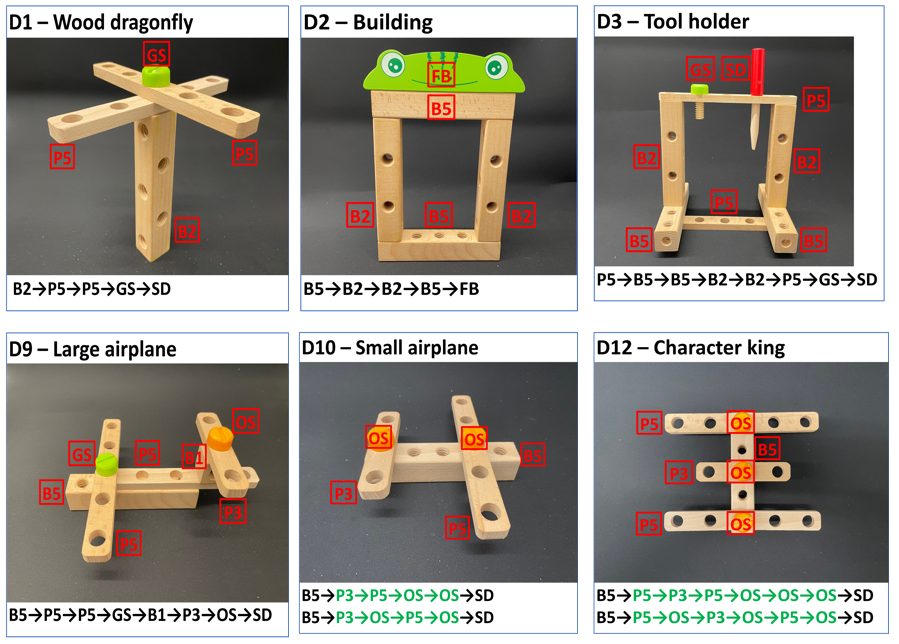}
	\caption{ Toy assembly tasks with one of instructions.}
	\label{assembly_tasks}
  \vspace{-0.8cm}
\end{figure}

\textbf{Experimental setup: }
We collected teleoperation sequences of human users performing assembly tasks by operating two robotic hands on a virtual reality setup in a simulation developed in ~\cite{belardinelli2022intention, manschitzshared}. 
The users performed $6$ assembly tasks in a virtual scene rendered via Rviz.
This was displayed in the HTC Vive Pro Eye headset, featuring
a $1440 \times 1600$ pixels screen per eye ($2880 \times 1600$ pixels combined, $110$ degrees of Field-Of-View), and a binocular Tobii
eyetracker working at $120$ Hz. The virtual
scene consisted of a table with $10$ types of toy assembly pieces. 

\textbf{Dataset collection: }
We collected data on $13$ participants performing $6$ assembly tasks (toys) that are shown in Fig.~\ref{assembly_tasks}.
In the figure, one of the instructions as a sequence of assembling pieces is displayed below the image. 
Except for inferring the assembly toys of the teleoperators, 
we are interested in capturing the natural order of actions in which the participants assemble the targeted toy.
We label actions based on their start and end times. There are $21$ actions in total that are designed based on the movements of end-effectors i.e., picking, placing, fastening, withdrawing, and associated objects. The actions span two or three stages i.e., pre-contact when the hand
(and tool) starts approaching the object, the interaction, and post-contact when the object is released.
This information from the dataset is crucial for intention estimation. Our dataset includes total of $202$ demonstrations of teleoperating $6$ tasks. During each process, we record 6D pose of objects in workspace and two-arm end-effectors, gaze direction, and video frames of egocentric views of teleoperators as shown in Fig.~\ref{demo} at 10Hz. The average demonstration duration is 1.5 minutes across the $6$ tasks. 
To test the improvement of hierarchical designs in diverse neural network structures, we split data into two types i.e., egocentric views and the rest features as the motion features. This helps to demonstrate the predictive capability of different feature combinations.

\textbf{Baselines: } We integrate our model with Graph Convolutional Network (GCN), LSTM, and SLOW-FAST neural networks that are mainly compared with two baselines: (1) Independent NN: applying these neural networks for task and action predictions independently, (2). NN-HMM: combining neural network models and  Hidden Markov Model~\cite{rabiner1986introduction} reasoning task and action in two stages of bottom-up manner.
Note that HMM itself is not able to take visual inputs as the SLOW-FAST model. We abbreviate 
Hierarchical as Hie in Table~\ref{tab:motion_acc}, \ref{tab:video_acc}, and \ref{tab:window_acc}.

\textbf{Training: } For each neural network model, we choose a data length of $35$ frames for task reasoning and $10$ frames for action prediction, corresponding to durations of 3.5 seconds and 1 second, respectively. The selection of these lengths aims to encompass sufficient information for both task and action.
The number of actions and tasks is not balanced. 
The model is always expected to perform well in the minority class as well as the majority class for multi-label classification.
Before training, we calculate the category weights of actions based on a balancing method, which adjusts weights inversely proportional to class frequencies, and then pass these weights into the optimizers at hierarchical levels to fit the model. Furthermore, we normalize the motion features to ensure they are treated equally by neural networks. 

\textbf{Evaluation: }  Accuracy is computed on a per-frame basis. The testing evaluation is real-time implementation.
The Slow-Fast model's network weights are initialized from the Kinetics-400 classification models. For a more in-depth understanding of the implementation, refer to~\cite{feichtenhofer2019slowfast}. In the case of NN-HMM, the model undergoes a two-stage bottom-up training process. Initially, LSTM and Slow-Fast serve as neural network models with diverse input types for training on the action estimation layer. Subsequently, the predicted actions are utilized as input data for HMM, where the Viterbi algorithm generates tasks.


\begin{table}[]
	\centering
	{
		\caption{The accuracy comparisons with data type of motion features, where Hierarchical is abbreviated as Hie. }
		\begin{tabular}{c|c|ccc}
			\hline
   Method | Accuracy & Data Type & Action  &  Task \\ \hline
			NN-HMM & motion & $92.27\%$ & $94.91\%$ \\ \hline 
			
			  LSTM & motion & $92.27\%$ &$96.79\%$ \\ \hline

             \textbf{Hie-LSTM} & motion & $\textbf{95.41\%}$ & $\textbf{98.25\%}$ \\ \hline

            GCN & motion & $89.98\%$ &$96.15\%$ \\ \hline

            \textbf{Hie-GCN} & motion & $\textbf{94.73\%}$ & $\textbf{97.10\%}$ \\ \hline

		\end{tabular}
		\label{tab:motion_acc}
	}
\end{table}

\begin{table}[]
	\centering
	{
		\caption{The accuracy comparisons with egocentric data, where Hierarchical is abbreviated as Hie.}
		\begin{tabular}{c|c|ccc}
			\hline
			
			  Method | Accuracy & Data Type & Action  &  Task \\ \hline
		    Slow-Fast & egocentric &$82.81\%$ & $84.57\%$ \\ \hline

                Slow-Fast-HMM & egocentric &$82.81\%$ & $85.94\%$ \\ \hline
      
			\textbf{Hie-Slow-Fast} & egocentric & \textbf{$\textbf{86.18\%}$} & $\textbf{87.33\%}$ \\ \hline
			
		\end{tabular}
		\label{tab:video_acc}
	}
\end{table}

\textbf{Results: } 
Initially, GCN and LSTM serve as fundamental models, handling motion features. The baseline (1) employs them separately for action and task intention estimation, overlooking hierarchy relations. The hierarchical HMM model also utilizes motion data as input, with improved accuracies demonstrated in Table~\ref{tab:motion_acc}. Hierarchical structures outperform alternative approaches.
In a second step, we integrate the hierarchical structure into Slow-Fast neural networks to showcase its generalization across diverse inputs. The resulting accuracy comparison is presented in Table~\ref{tab:video_acc}, indicating enhanced performance of video models with the hierarchical structure.

\begin{figure}
    \centering
    	\subfloat[Normalized confusion matrix of task prediction using hierarchical LSTM (Left) and hierarchical Slow-Fast (Right).]{\includegraphics[width=0.95\linewidth]{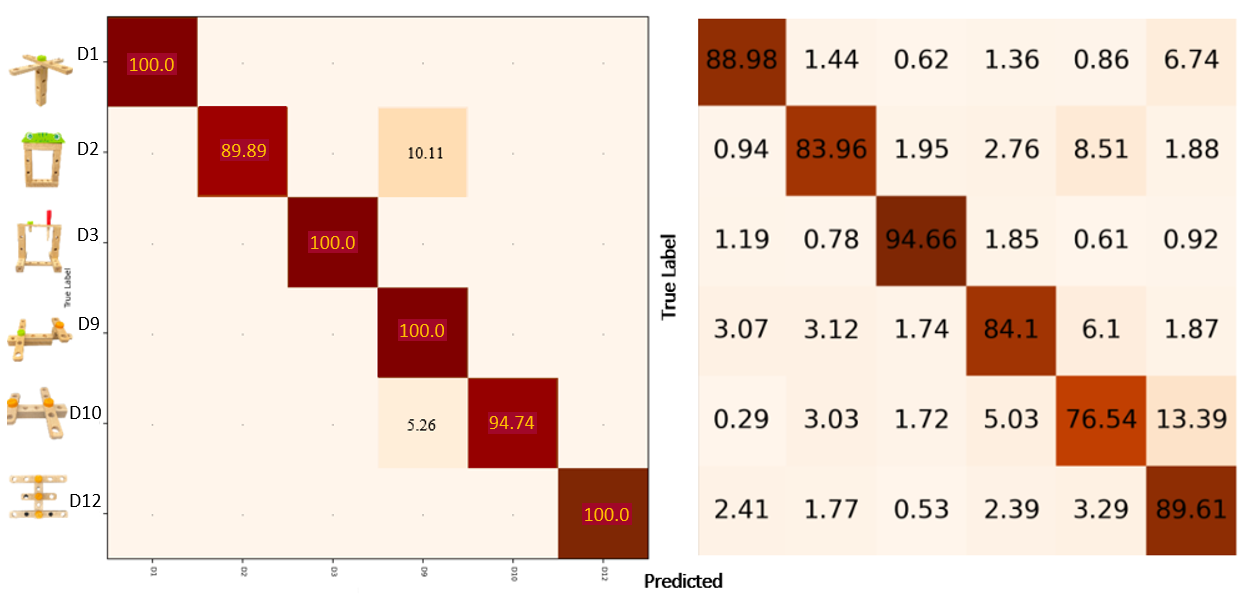}\label{Task_accuracy}} \\
     
	\subfloat[Normalized confusion matrix of action prediction using hierarchical Slow-Fast]{\includegraphics[width=0.75\linewidth]{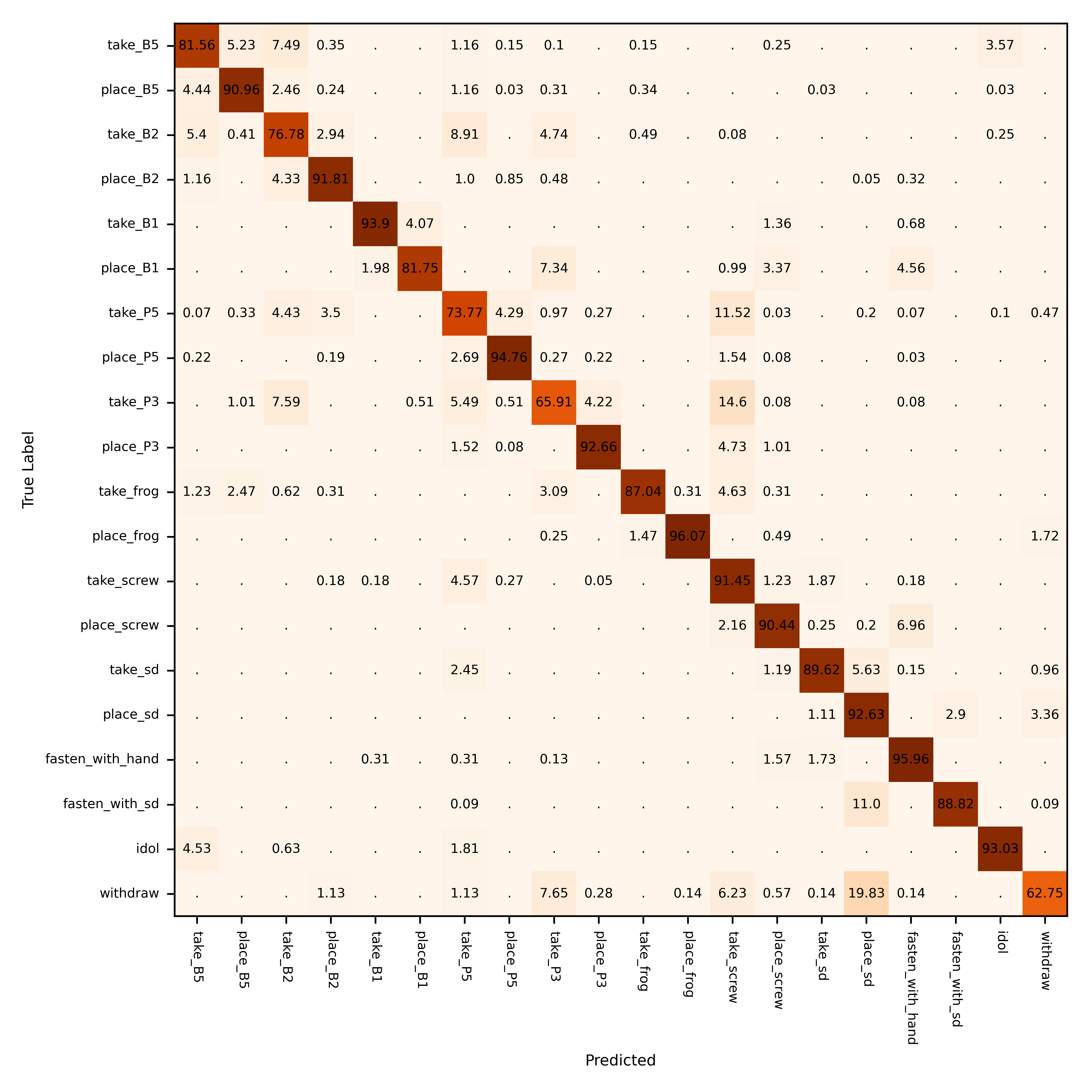}\label{SF_action_accuracy}} \\
        \subfloat[Normalized confusion matrix of action prediction using hierarchical LSTM]{\includegraphics[width=0.75\linewidth]{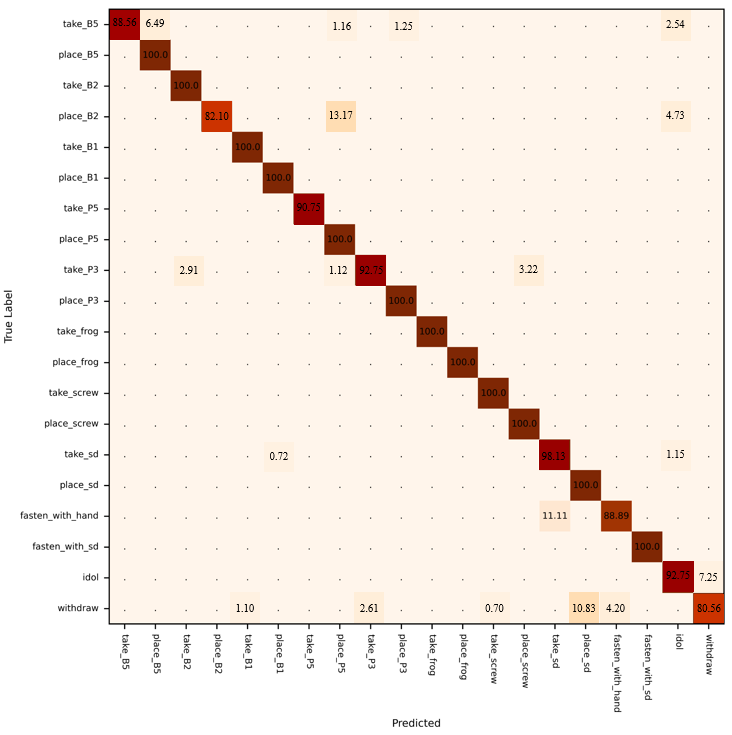}\label{LSTM_action_accuracy}}
    \caption{On the x axis the predictions, on the
y the ground truth. Numbers represents the frequency with which samples
of a certain class (row) were classified with the label on the corresponding}
    \label{action_accuracy}
\end{figure}

The confusion matrix in Fig.\ref{action_accuracy} depicts the performance of using LSTM and Slow-Fast as backbones for the hierarchical model. Examining Fig.\ref{Task_accuracy}, mispredictions between tasks D$9$ and D$10$ are noticeable due to shared configurations at the beginning. Despite this, at least $50\%$ accuracy in recognizing intentions for each action is achieved on average, dependent on the uniqueness of movements and objects.

Moreover, the significance of employing a multi-window strategy is demonstrated in section~\ref{sub:mask}. Our complete framework, denoted as Hierarchical NN-W and Hierarchical SF-W, outperforms baselines (Hierarchical NN-O and Hierarchical SF-O) without the multi-window strategy, as shown in Table~\ref{tab:window_acc}. Utilizing more informative data inputs improves accuracies, eliminating the need for the neural network to extract features. Future work will explore an auto-tuning method for optimizing window sizes.





 \begin{figure}
	\centering 	\includegraphics[width=0.36\textheight]{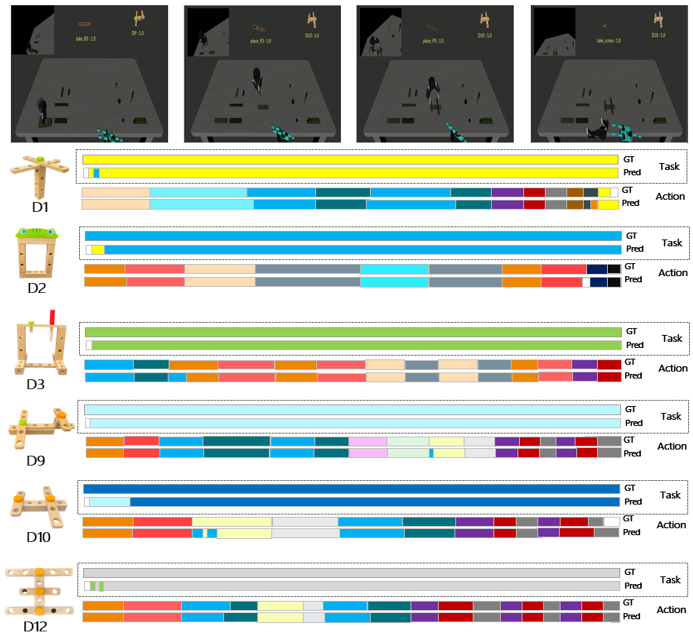}
	\caption{We select $6$ videos from each assembly task. For each video, we show the ground-truth task label, predictions results.}
	\label{online_demo}
\end{figure}

\begin{table}[]
\centering

\vspace{-0.5cm}
{
    \caption{The results of accuracy comparisons on different data types, where Hierarchical is abbreviated as Hie. }
    \begin{tabular}{c|c|ccc}
        \hline
        
        Method | Accuracy & Data Type & Action  &  Task \\ \hline
        Hie-SF-O & egocentric &$82.21\%$ & $84.57\%$ \\ \hline
        \textbf{Hie-SF-W} & egocentric & \textbf{$\textbf{86.18\%}$} & $\textbf{87.33\%}$ \\ \hline

        Hie-NN-O & motion & $93.82\%$ &$95.89\%$ \\ \hline

        \textbf{Hie- NN-W} & motion & $\textbf{95.41\%}$ & $\textbf{98.25\%}$ \\ \hline
        
    \end{tabular}
    \label{tab:window_acc}
}
\end{table}

Finally, we apply the LSTM and motion data to conduct qualitative task and action prediction results of our hierarchical deep learning models on teleoperation videos recorded in ROS bags shown in Fig.~\ref{online_demo}. 
The intention estimation produces results efficiently with $2$ Hz. The ground truth and prediction results are shown below the image sequence. Each pair of frames corresponds to the time at each second. By comparing the results of tasks $D9$ and $D10$, our model can sensitively recognize tasks based on the differences in sequential actions that appeared at early stages.

\section{Discussion and Conclusion} 

In this analysis, we find that incorporating hierarchical relations into intention estimation systems enhances prediction performance. This structure encourages top-down layers to share prediction information through hierarchical relations, expediting accurate predictions. Our online testing reveals that high-level intention abstraction increases the stability of low-level inference, reducing switches and improving robot control stability.
To summarize, we propose hierarchical deep learning models that consider dependent relations among multi-layer intentions and enforce their hierarchical structure during training. Our model is adaptable for integration with existing neural networks. For hierarchical data input, we introduce a multi-window strategy to mask unnecessary information at each layer, resulting in diverse inputs and embeddings.
We demonstrate enhanced inference performance with both motion and vision data compared to independent neural network models in real-time teleoperation Vive systems. In future research, we will explore the intention estimation model's robustness to teleoperation anomalies and the generalization capability of foundation models using zero-shot or few-shot learning.

\section{Acknowledgment} 
We would like to express our deep and sincere gratitude to Dirk Ruiken and Simon Manschitz for providing the robotic simulation environment and their kind support throughout the project. Also we would like to thank our colleagues for helping us with the data collection and valuable comments.

\bibliographystyle{IEEEtran}
\bibliography{reference}

\begin{thebibliography}{10}
\providecommand{\url}[1]{#1}
\csname url@samestyle\endcsname
\providecommand{\newblock}{\relax}
\providecommand{\bibinfo}[2]{#2}
\providecommand{\BIBentrySTDinterwordspacing}{\spaceskip=0pt\relax}
\providecommand{\BIBentryALTinterwordstretchfactor}{4}
\providecommand{\BIBentryALTinterwordspacing}{\spaceskip=\fontdimen2\font plus
\BIBentryALTinterwordstretchfactor\fontdimen3\font minus
  \fontdimen4\font\relax}
\providecommand{\BIBforeignlanguage}[2]{{%
\expandafter\ifx\csname l@#1\endcsname\relax
\typeout{** WARNING: IEEEtran.bst: No hyphenation pattern has been}%
\typeout{** loaded for the language `#1'. Using the pattern for}%
\typeout{** the default language instead.}%
\else
\language=\csname l@#1\endcsname
\fi
#2}}
\providecommand{\BIBdecl}{\relax}
\BIBdecl

\bibitem{alevizos2020physical}
K.~I. Alevizos, C.~P. Bechlioulis, and K.~J. Kyriakopoulos, ``Physical
  human--robot cooperation based on robust motion intention estimation,''
  \emph{Robotica}, vol.~38, no.~10, pp. 1842--1866, 2020.

\bibitem{fang2023human}
C.~Fang, L.~Peternel, A.~Seth, M.~Sartori, K.~Mombaur, and E.~Yoshida, ``Human
  modeling in physical human-robot interaction: A brief survey,'' \emph{IEEE
  Robotics and Automation Letters}, 2023.

\bibitem{wilcox2013optimization}
R.~Wilcox, S.~Nikolaidis, and J.~Shah, ``Optimization of temporal dynamics for
  adaptive human-robot interaction in assembly manufacturing,''
  \emph{Robotics}, vol.~8, no. 441, pp. 10--15, 2013.

\bibitem{zhou2023local}
Z.~Zhou, S.~Wang, Z.~Chen, M.~Cai, H.~Wang, Z.~Li, and Z.~Kan, ``Local
  observation based reactive temporal logic planning of human-robot systems,''
  \emph{IEEE Transactions on Automation Science and Engineering}, 2023.

\bibitem{li2022assimilation}
G.~Li, Z.~Li, and Z.~Kan, ``Assimilation control of a robotic exoskeleton for
  physical human-robot interaction,'' \emph{IEEE Robotics and Automation
  Letters}, vol.~7, no.~2, pp. 2977--2984, 2022.

\bibitem{schydlo2018anticipation}
P.~Schydlo, M.~Rakovic, L.~Jamone, and J.~Santos-Victor, ``Anticipation in
  human-robot cooperation: A recurrent neural network approach for multiple
  action sequences prediction,'' in \emph{2018 IEEE International Conference on
  Robotics and Automation (ICRA)}.\hskip 1em plus 0.5em minus 0.4em\relax IEEE,
  2018, pp. 5909--5914.

\bibitem{nicolis2018human}
D.~Nicolis, A.~M. Zanchettin, and P.~Rocco, ``Human intention estimation based
  on neural networks for enhanced collaboration with robots,'' in \emph{2018
  IEEE/RSJ International Conference on Intelligent Robots and Systems
  (IROS)}.\hskip 1em plus 0.5em minus 0.4em\relax IEEE, 2018, pp. 1326--1333.

\bibitem{belardinelli2022intention}
A.~Belardinelli, A.~R. Kondapally, D.~Ruiken, D.~Tanneberg, and T.~Watabe,
  ``Intention estimation from gaze and motion features for human-robot
  shared-control object manipulation,'' in \emph{2022 IEEE/RSJ International
  Conference on Intelligent Robots and Systems (IROS)}.\hskip 1em plus 0.5em
  minus 0.4em\relax IEEE, 2022, pp. 9806--9813.

\bibitem{huang2023hierarchical}
Z.~Huang, Y.-J. Mun, X.~Li, Y.~Xie, N.~Zhong, W.~Liang, J.~Geng, T.~Chen, and
  K.~Driggs-Campbell, ``Hierarchical intention tracking for robust human-robot
  collaboration in industrial assembly tasks,'' in \emph{2023 IEEE
  International Conference on Robotics and Automation (ICRA)}.\hskip 1em plus
  0.5em minus 0.4em\relax IEEE, 2023, pp. 9821--9828.

\bibitem{manschitzshared}
S.~Manschitz and D.~Ruiken, ``Shared autonomy for intuitive teleoperation,''
  \emph{ICRA Workshop: Shared Autonomy in Physical Human-Robot Interaction:
  Adaptability and Trust}, May 2022.

\bibitem{gao2020deep}
D.~Gao, W.~Yang, H.~Zhou, Y.~Wei, Y.~Hu, and H.~Wang, ``Deep hierarchical
  classification for category prediction in e-commerce system,'' \emph{arXiv
  preprint arXiv:2005.06692}, 2020.

\bibitem{sheridan1995teleoperation}
T.~B. Sheridan, ``Teleoperation, telerobotics and telepresence: A progress
  report,'' \emph{Control Engineering Practice}, vol.~3, no.~2, pp. 205--214,
  1995.

\bibitem{rozo2016learning}
L.~Rozo, S.~Calinon, D.~G. Caldwell, P.~Jimenez, and C.~Torras, ``Learning
  physical collaborative robot behaviors from human demonstrations,''
  \emph{IEEE Transactions on Robotics}, vol.~32, no.~3, pp. 513--527, 2016.

\bibitem{yu2005telemanipulation}
W.~Yu, R.~Alqasemi, R.~Dubey, and N.~Pernalete, ``Telemanipulation assistance
  based on motion intention recognition,'' in \emph{Proceedings of the 2005
  IEEE international conference on robotics and automation}.\hskip 1em plus
  0.5em minus 0.4em\relax IEEE, 2005, pp. 1121--1126.

\bibitem{dragan2013policy}
A.~D. Dragan and S.~S. Srinivasa, ``A policy-blending formalism for shared
  control,'' \emph{The International Journal of Robotics Research}, vol.~32,
  no.~7, pp. 790--805, 2013.

\bibitem{hauser2013recognition}
K.~Hauser, ``Recognition, prediction, and planning for assisted teleoperation
  of freeform tasks,'' \emph{Autonomous Robots}, vol.~35, pp. 241--254, 2013.

\bibitem{aarno2008motion}
D.~Aarno and D.~Kragic, ``Motion intention recognition in robot assisted
  applications,'' \emph{Robotics and Autonomous Systems}, vol.~56, no.~8, pp.
  692--705, 2008.

\bibitem{tanwani2017generative}
A.~K. Tanwani and S.~Calinon, ``A generative model for intention recognition
  and manipulation assistance in teleoperation,'' in \emph{2017 IEEE/RSJ
  International Conference on Intelligent Robots and Systems (IROS)}.\hskip 1em
  plus 0.5em minus 0.4em\relax IEEE, 2017, pp. 43--50.

\bibitem{rabiner1989tutorial}
L.~R. Rabiner, ``A tutorial on hidden markov models and selected applications
  in speech recognition,'' \emph{Proceedings of the IEEE}, vol.~77, no.~2, pp.
  257--286, 1989.

\bibitem{tanwani2016learning}
A.~K. Tanwani and S.~Calinon, ``Learning robot manipulation tasks with
  task-parameterized semitied hidden semi-markov model,'' \emph{IEEE Robotics
  and Automation Letters}, vol.~1, no.~1, pp. 235--242, 2016.

\bibitem{zhu2008human}
C.~Zhu, Q.~Cheng, and W.~Sheng, ``Human intention recognition in smart assisted
  living systems using a hierarchical hidden markov model,'' in \emph{2008 IEEE
  International Conference on Automation Science and Engineering}.\hskip 1em
  plus 0.5em minus 0.4em\relax IEEE, 2008, pp. 253--258.

\bibitem{li2013human}
Y.~Li and S.~S. Ge, ``Human--robot collaboration based on motion intention
  estimation,'' \emph{IEEE/ASME Transactions on Mechatronics}, vol.~19, no.~3,
  pp. 1007--1014, 2013.

\bibitem{de2016neural}
E.~De~Momi, L.~Kranendonk, M.~Valenti, N.~Enayati, and G.~Ferrigno, ``A neural
  network-based approach for trajectory planning in robot--human handover
  tasks,'' \emph{Frontiers in Robotics and AI}, vol.~3, p.~34, 2016.

\bibitem{lea2016learning}
C.~Lea, R.~Vidal, and G.~D. Hager, ``Learning convolutional action primitives
  for fine-grained action recognition,'' in \emph{2016 IEEE international
  conference on robotics and automation (ICRA)}.\hskip 1em plus 0.5em minus
  0.4em\relax IEEE, 2016, pp. 1642--1649.

\bibitem{yuan2022leveraging}
C.~Yuan, T.~Marion, and M.~Moghaddam, ``Leveraging end-user data for enhanced
  design concept evaluation: A multimodal deep regression model,''
  \emph{Journal of Mechanical Design}, vol. 144, no.~2, p. 021403, 2022.

\bibitem{lu2020human}
W.~Lu, Z.~Hu, and J.~Pan, ``Human-robot collaboration using variable admittance
  control and human intention prediction,'' in \emph{2020 IEEE 16th
  International Conference on Automation Science and Engineering (CASE)}.\hskip
  1em plus 0.5em minus 0.4em\relax IEEE, 2020, pp. 1116--1121.

\bibitem{bandouch2009tracking}
J.~Bandouch and M.~Beetz, ``Tracking humans interacting with the environment
  using efficient hierarchical sampling and layered observation models,'' in
  \emph{2009 IEEE 12th International Conference on Computer Vision Workshops,
  ICCV Workshops}.\hskip 1em plus 0.5em minus 0.4em\relax IEEE, 2009, pp.
  2040--2047.

\bibitem{hayes2016autonomously}
B.~Hayes and B.~Scassellati, ``Autonomously constructing hierarchical task
  networks for planning and human-robot collaboration,'' in \emph{2016 IEEE
  International Conference on Robotics and Automation (ICRA)}.\hskip 1em plus
  0.5em minus 0.4em\relax IEEE, 2016, pp. 5469--5476.

\bibitem{holtzen2016inferring}
S.~Holtzen, Y.~Zhao, T.~Gao, J.~B. Tenenbaum, and S.-C. Zhu, ``Inferring human
  intent from video by sampling hierarchical plans,'' in \emph{2016 IEEE/RSJ
  International Conference on Intelligent Robots and Systems (IROS)}.\hskip 1em
  plus 0.5em minus 0.4em\relax IEEE, 2016, pp. 1489--1496.

\bibitem{han2016interactive}
J.-H. Han, S.-H. Choi, and J.-H. Kim, ``Interactive human intention reading by
  learning hierarchical behavior knowledge networks for human-robot
  interaction,'' \emph{ETRI Journal}, vol.~38, no.~6, pp. 1229--1239, 2016.

\bibitem{feichtenhofer2019slowfast}
C.~Feichtenhofer, H.~Fan, J.~Malik, and K.~He, ``Slowfast networks for video
  recognition,'' in \emph{Proceedings of the IEEE/CVF international conference
  on computer vision}, 2019, pp. 6202--6211.

\bibitem{medsker2001recurrent}
L.~R. Medsker and L.~Jain, ``Recurrent neural networks,'' \emph{Design and
  Applications}, vol.~5, no. 64-67, p.~2, 2001.

\bibitem{graves2012long}
A.~Graves and A.~Graves, ``Long short-term memory,'' \emph{Supervised sequence
  labelling with recurrent neural networks}, pp. 37--45, 2012.

\bibitem{vaswani2017attention}
A.~Vaswani, N.~Shazeer, N.~Parmar, J.~Uszkoreit, L.~Jones, A.~N. Gomez,
  {\L}.~Kaiser, and I.~Polosukhin, ``Attention is all you need,''
  \emph{Advances in neural information processing systems}, vol.~30, 2017.

\bibitem{abi2017learning}
F.~Abi-Farraj, T.~Osa, N.~P.~J. Peters, G.~Neumann, and P.~R. Giordano, ``A
  learning-based shared control architecture for interactive task execution,''
  in \emph{2017 IEEE international conference on robotics and automation
  (ICRA)}.\hskip 1em plus 0.5em minus 0.4em\relax IEEE, 2017, pp. 329--335.

\bibitem{feddema1989vision}
J.~T. Feddema and O.~R. Mitchell, ``Vision-guided servoing with feature-based
  trajectory generation (for robots),'' \emph{IEEE Transactions on Robotics and
  Automation}, vol.~5, no.~5, pp. 691--700, 1989.

\bibitem{marcucci2022motion}
T.~Marcucci, M.~Petersen, D.~von Wrangel, and R.~Tedrake, ``Motion planning
  around obstacles with convex optimization,'' \emph{arXiv preprint
  arXiv:2205.04422}, 2022.

\bibitem{shi2023waypoint}
L.~X. Shi, A.~Sharma, T.~Z. Zhao, and C.~Finn, ``Waypoint-based imitation
  learning for robotic manipulation,'' \emph{arXiv preprint arXiv:2307.14326},
  2023.

\bibitem{rabiner1986introduction}
L.~Rabiner and B.~Juang, ``An introduction to hidden markov models,''
  \emph{ieee assp magazine}, vol.~3, no.~1, pp. 4--16, 1986.

\end{thebibliography}

\end{document}